\documentclass{article}

\usepackage[preprint]{corl_2026}

\usepackage{graphicx}
\usepackage{booktabs}
\usepackage{amsmath,amssymb}
\graphicspath{{figures/}}

\newcommand{\codeavailability}{Code is available at \url{https://github.com/eqcy/FineCraft}.}

\title{FineCraft: Enhancing VLA Models for Fine-Grained Manipulation}
\author{
  Shuaike Zhang\textsuperscript{1} \quad Ziheng Quan\textsuperscript{2} \quad Junhao Ran\textsuperscript{1}\\
  \textbf{Muhe Ding\textsuperscript{1} \quad Shaokun Wang\textsuperscript{1} \quad Jianlong Wu\textsuperscript{1,3}}\\
  \textsuperscript{1} Harbin Institute of Technology, Shenzhen\\
  \textsuperscript{2} Xi'an Jiaotong University \quad \textsuperscript{3} Shenzhen Loop Area Institute\\
  HITSZ Robotics
}
\hypersetup{
  pdftitle={FineCraft: Enhancing VLA Models for Fine-Grained Manipulation},
  pdfauthor={Shuaike Zhang; Ziheng Quan; Junhao Ran; Muhe Ding; Shaokun Wang; Jianlong Wu},
  pdfsubject={MetaFine Challenge Technical Report -- HITSZ Robotics -- Preprint},
  pdfkeywords={Fine-Grained Manipulation, Vision-Language-Action Models, Robot Learning}
}

\newif\ifreportreferences
\reportreferencestrue

\begin{document}
\maketitle

\begin{abstract}
Fine-grained manipulation couples instruction grounding, local visual precision, and sustained execution. We present FineCraft, a framework that adapts pretrained models for fine-grained manipulation in the MetaFine Challenge. A task-routed mixture of policies selects independently adapted DINOv2-DiT or pretrained $\pi_{0.5}$ policies through a common observation-action interface. Local visual features help identify the instructed object parts, demonstrations and expert continuations teach the required actions, and new observations guide replanning during execution. FineCraft ranks fourth with an overall score of $0.878$, exceeding the published reference on all five tasks. Official results and local diagnostics characterize performance across visual perturbations, instruction changes, and multi-stage manipulation.
\codeavailability

\end{abstract}


\section{Introduction}
\label{sec:introduction}

Fine-grained manipulation requires acting on the instructed part or attribute while respecting spatial and temporal constraints. MetaFine~\citep{xu2026beyond} evaluates understanding, perception, and controlled behavior, distinguishing correct target selection from sustained execution through a longer manipulation sequence.

Pretrained robot policies transfer visual and motor knowledge to new tasks~\citep{octo2024,kim2025openvla}. Instruction-conditioned affordance reasoning~\citep{tang2025affordgrasp}, task-specific adapters for class-incremental vision~\citep{li2025dia}, and reusable skill composition~\citep{zhang2026sce} motivate adapting representation, supervision, and execution to fine-grained requirements.

FineCraft organizes DINOv2-DiT and $\pi_{0.5}$ backbones as a task-routed mixture of independently adapted policies. We adapt visual features and training demonstrations to each task's target objects and instructions, and replan actions using new observations during execution. We evaluate all five tasks through official capability profiles and local grasp and switch-adaptation diagnostics.

\section{Method}
\label{sec:method}

\begin{figure}[t]
  \centering
  \includegraphics[width=\linewidth]{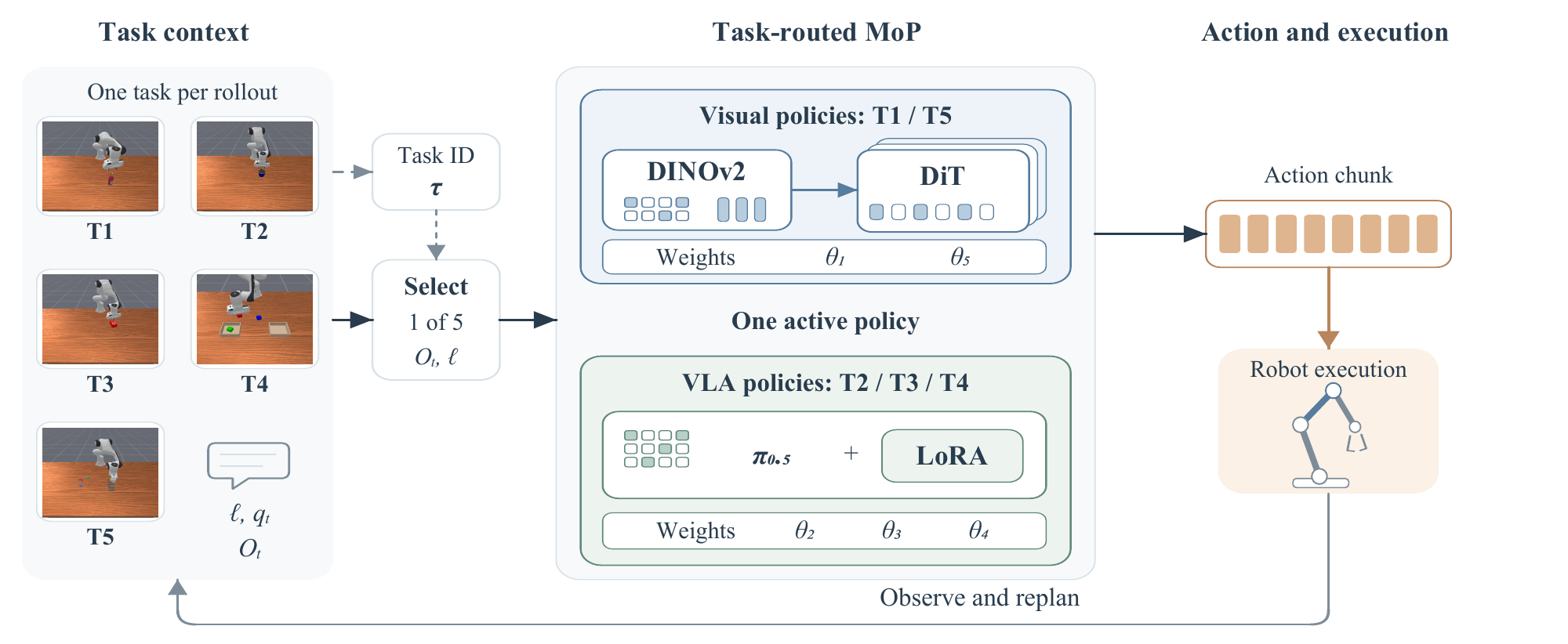}
  \caption{FineCraft as a task-routed mixture of policies (MoP). The task ID selects one independent task policy for the rollout. Two backbone families summarize five task instances, including their task-specific branches. The selected policy generates one action stream and replans from new observations.}
  \label{fig:framework}
\end{figure}

\paragraph{Task selection and action execution.}
At episode reset, the known task ID $\tau\in\{1,\ldots,5\}$ selects one independent task policy. The observation $o_t=(I_t^{\mathrm{base}},I_t^{\mathrm{wrist}},q_t)$ consists of two $512\times512$ RGB images and the proprioceptive state $q_t$, represented by nine joint positions. Let $\ell$ denote the task instruction and $\mathcal{O}_t$ denote the recent observation history, augmented with previous actions where required by the policy. The task-routed mixture is
\begin{equation}
  \pi_{\mathrm{MoP}}(\mathbf{A}_t\mid\mathcal{O}_t,\ell,\tau)
  =\sum_{k=1}^{5}\mathbf{1}[\tau=k]\,
    \pi_{\theta_k}(\mathbf{A}_t\mid\mathcal{O}_t,\ell).
  \label{eq:policy}
\end{equation}
Actions $a_t\in\mathbb{R}^{8}$ control seven arm joints and one gripper. Each prediction contains $H$ actions, $\mathbf{A}_t=(a_t,\ldots,a_{t+H-1})$, of which $K\leq H$ are executed before observing and replanning~\citep{zhao2023act,chi2023diffusion}. T2/T4 use $H=K=10$; T3 uses $H=10$, $K=5$.

\paragraph{From images to action sequences.}
For bottle-part grasping and letter insertion (T1/T5), DINOv2~\citep{oquab2023dinov2} extracts spatial feature maps from both views. Trained heatmap and visibility heads convert these maps into image keypoints and confidence values. The instruction selects the cap/body branch in T1 or the target letter in T5. The action model uses a two-frame observation history containing predicted visual features, joint positions, and previous actions, together with a one-hot indicator of the target. For cap grasping, we also pool local visual features around the predicted cap center and gripper. For insertion, the visual features describe the requested letter and its slot, together with the surrounding objects.

A diffusion Transformer (DiT)~\citep{peebles2022dit} predicts actions from these features. During training, noise is added to demonstration action sequences and the network learns to predict that noise. At inference, denoising converts Gaussian noise into actions. The T1 branches and the T5 general policy predict 16 actions and execute two before replanning. Landmark and visibility labels supervise perception; action prediction uses the model's own visual estimates.

\paragraph{Adapting pretrained robot policies.}
For mug transport, switch toggling, and cube sorting (T2, T3, and T4), we fine-tune $\pi_{0.5}$~\citep{pi2025pi05} on task demonstrations. Images, joint positions, and the instruction enter its pretrained vision-language and action modules. LoRA~\citep{hu2021lora} adapts the Transformer weights through
\begin{equation}
  W=W_0+UV^{\top},\qquad
  U\in\mathbb{R}^{d_{\mathrm{out}}\times r},\quad
  V\in\mathbb{R}^{d_{\mathrm{in}}\times r},
  \label{eq:lora}
\end{equation}
where $W_0\in\mathbb{R}^{d_{\mathrm{out}}\times d_{\mathrm{in}}}$ is the pretrained weight matrix and $r$ is the rank of the update. Each task has separate weights and action-normalization statistics. T2 and T4 each use $300$ demonstrations and $30{,}000$ updates, learning the instructed transport direction or sequence of cube placements.

\paragraph{Learning the instructed contacts.}
T1 uses separate cap and body DINOv2-DiT models. The cap model is trained on complete successful trajectories generated offline by a $\pi_{0.5}$ teacher, covering approach, alignment, and gripper closure. For the body model, an offline expert continues from difficult states reached by the learned policy; successful expert actions are mixed with original demonstrations~\citep{ross2011dagger}. Learner actions before intervention are excluded from expert-action targets, and supervised windows end at interruptions in expert control. During replanning, three candidate chunks are compared with the unexecuted arm actions of the previous chunk. The lowest-discrepancy chunk is selected when its normalized cost is within a threshold; otherwise, the first sample is used. The gripper command comes from the selected complete chunk.

For T3, red and blue commands are demonstrated in the same physical scenes, so changing the instruction requires changing the selected switch. Additional scenes cover camera and lighting perturbations. The resulting $864$ trajectories contain $88{,}059$ action frames, including approach, toggling, and holding. Training and inference use the same explicit target-color prompt. LoRA training retains $\pi_{0.5}$'s original objective for learning action sequences from demonstrations.

\paragraph{Completing insertion with feedback.}
T5 combines a general DiT policy, a finishing policy that predicts sequences of eight actions, and letter-specific correction policies. A four-frame recurrent encoder combines predicted letter/slot keypoints, their relative image positions, local visual features, robot/action history, and elapsed time. A learned phase classifier switches from the general policy to the finishing policy after two consecutive positive predictions. A second classifier monitors recent feature changes; persistent predicted stalls can activate a correction policy for C, o, or L. For C and o, a further classifier suppresses correction when the current policy is likely to complete insertion. Stage changes are one-way, from general to finishing to correction. The short-horizon policies learn from successful offline expert continuations and replayed successes while the general policy remains fixed. Each policy executes two actions before receiving a new observation.

\section{Experiments}
\label{sec:experiments}

\subsection{Evaluation protocol}
\label{sec:protocol}

We follow the five-task challenge protocol~\citep{metafine_challenge} and MetaFine's~\citep{xu2026beyond} perception ($P$), understanding ($U$), and behavior ($B$) axes. Perception averages normalized areas under camera- and lighting-perturbation success curves. Understanding varies instructions over fixed scene seeds. Behavior measures final success for single-stage tasks and normalized stage progress for multi-stage tasks. T1/T3/T5 reuse $U$ outcomes for $B$. Scores are
\begin{equation}
  S_\tau = \frac{P_\tau+U_\tau+B_\tau}{3}, \qquad
  S_{\mathrm{overall}} = \frac{1}{5}\sum_{\tau=1}^{5} S_\tau.
  \label{eq:score}
\end{equation}
Scores lie in $[0,1]$ and are shown to three decimals; differences are expressed in percentage points (pp). We use the organizer's final per-task entries for HITSZ Robotics from the public leaderboard updated on October 24, 2026~\citep{metafine_leaderboard}. The organizer's published $\pi_0$~\citep{black2024pi0} scores provide a reference across tasks. Local evaluations of T1 and T3 use the released evaluator.

\begin{table}[ht]
\centering
\caption{Official five-task results for FineCraft, listed as HITSZ Robotics on the final leaderboard~\citep{metafine_leaderboard}. Scores use the aggregation in Equation~\ref{eq:score}.}
\label{tab:official}
\begin{tabular}{llrrrr}
\toprule
Task & Goal & $P$ & $U$ & $B$ & Score \\
\midrule
T1 & Bottle-part grasping & 0.787 & 0.800 & 0.800 & 0.796 \\
T2 & Mug transport & 1.000 & 1.000 & 1.000 & 1.000 \\
T3 & Switch toggling & 0.925 & 0.925 & 0.925 & 0.925 \\
T4 & Cube sorting & 0.995 & 0.983 & 0.983 & 0.987 \\
T5 & Letter insertion & 0.594 & 0.725 & 0.725 & 0.681 \\
\midrule
\multicolumn{2}{l}{Overall (rank 4)} & 0.860 & 0.887 & 0.887 & 0.878 \\
\bottomrule
\end{tabular}
\end{table}

\subsection{Official capability profiles}
\paragraph{T2 and T4: multi-stage execution.}
Mug transport reaches $1.000$ in all dimensions. Cube sorting scores $0.987$, with behavior at $0.983$ measuring progress through three placements. These results cover instructed transport and repeated placement beyond the initial grasp.

\begin{figure}[t]
  \centering
  \includegraphics[width=\linewidth]{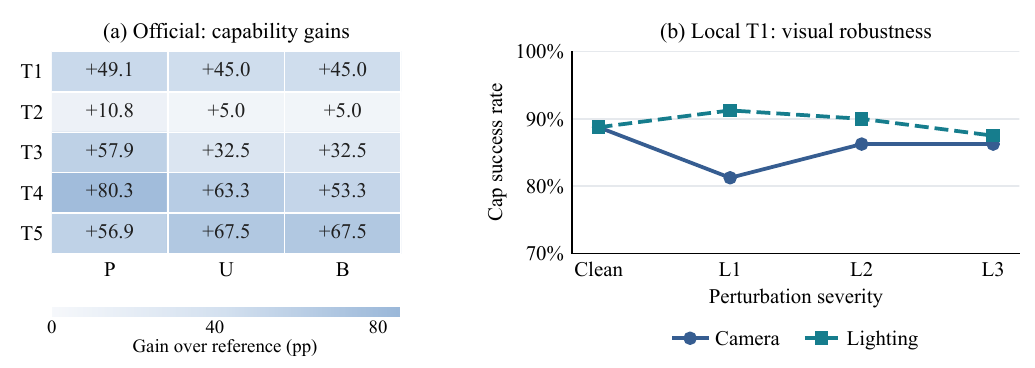}
  \caption{Capability gains and local perception diagnostics. (a) Official improvements in $P$, $U$, and $B$ over the published $\pi_0$ reference, in percentage points (pp)~\citep{metafine_leaderboard}. (b) Local T1 cap success under camera and lighting perturbations, with $80$ rollouts per condition.}
  \label{fig:comparison}
\end{figure}

\paragraph{Comparison with the organizer reference.}
FineCraft outperforms the published $\pi_0$ reference on every task and evaluation dimension (Figure~\ref{fig:comparison}a). The overall gain is $44.8$ percentage points over its mean of $0.430$. Cube sorting has the largest perception gain, while letter insertion has the largest instruction-conditioned gain.

\subsection{Diagnostic analysis}
\label{sec:task_analysis}
\paragraph{T1: grasping the instructed part.}
In local tests of the updated cap/body configuration, body and cap commands succeed in $70/80$ and $72/80$ episodes over the same ordered scene seeds. The cap's camera and lighting tests (Figure~\ref{fig:comparison}b) add $560$ rollouts, yielding perception $0.874$ and a composite score of $0.883$ across $720$ rollouts. The area under the success curve is smaller for camera perturbations than for lighting changes.

\paragraph{T5: insertion.}
On T5, FineCraft improves the official perception and understanding scores over the reference by $56.9$ and $67.5$ percentage points, respectively. Its lower perception score indicates that camera and lighting perturbations remain more challenging than instruction changes in fixed scenes.

\paragraph{T3: a compact adaptation study.}
Across three local development groups ($60$ scenes and $540$ rollouts per model), continued training increases the mean evaluation score from $0.841$ to $0.894$, with improvements in all three groups. Training and evaluation scene seeds are disjoint. This study evaluates the combined effect of expanded training data and continued policy adaptation.

\section{Conclusion}
\label{sec:conclusion}
FineCraft adapts visual diffusion policies and VLA models to individual tasks using demonstrations and replans actions from updated observations during execution. The HITSZ Robotics entry ranks fourth in MetaFine and exceeds the published reference on all five tasks. Official capability profiles and local diagnostics characterize visual robustness, instructed contacts, and multi-stage execution.

\ifreportreferences
  \clearpage
  \bibliography{example}

\begin{thebibliography}{16}
\providecommand{\natexlab}[1]{#1}
\providecommand{\url}[1]{\texttt{#1}}
\expandafter\ifx\csname urlstyle\endcsname\relax
  \providecommand{\doi}[1]{doi: #1}\else
  \providecommand{\doi}{doi: \begingroup \urlstyle{rm}\Url}\fi

\bibitem[Xu et~al.(2026)Xu, Zhang, Ge, Hao, Belongie, Geng, Peng, and
  Wei]{xu2026beyond}
H.-Y. Xu, P.~Zhang, Z.~Ge, X.~Hao, S.~Belongie, X.~Geng, Y.~Peng, and X.-S.
  Wei.
\newblock Beyond binary success: A diagnostic meta-evaluation framework for
  fine-grained manipulation.
\newblock \emph{arXiv preprint arXiv:2605.19986}, 2026.

\bibitem[Ghosh et~al.(2024)Ghosh, Walke, Pertsch, Black, Mees, Dasari, Hejna,
  Kreiman, Xu, Luo, Tan, Chen, Vuong, Xiao, Sanketi, Sadigh, Finn, and
  Levine]{octo2024}
D.~Ghosh, H.~R. Walke, K.~Pertsch, K.~Black, O.~Mees, S.~Dasari, J.~Hejna,
  T.~Kreiman, C.~Xu, J.~Luo, Y.~L. Tan, L.~Y. Chen, Q.~Vuong, T.~Xiao, P.~R.
  Sanketi, D.~Sadigh, C.~Finn, and S.~Levine.
\newblock {Octo}: An open-source generalist robot policy.
\newblock In \emph{Proceedings of Robotics: Science and Systems}, 2024.

\bibitem[Kim et~al.(2025)Kim, Pertsch, Karamcheti, Xiao, Balakrishna, Nair,
  Rafailov, Foster, Sanketi, Vuong, Kollar, Burchfiel, Tedrake, Sadigh, Levine,
  Liang, and Finn]{kim2025openvla}
M.~J. Kim, K.~Pertsch, S.~Karamcheti, T.~Xiao, A.~Balakrishna, S.~Nair,
  R.~Rafailov, E.~P. Foster, P.~R. Sanketi, Q.~Vuong, T.~Kollar, B.~Burchfiel,
  R.~Tedrake, D.~Sadigh, S.~Levine, P.~Liang, and C.~Finn.
\newblock {OpenVLA}: An open-source vision-language-action model.
\newblock In \emph{Proceedings of The 8th Conference on Robot Learning}, volume
  270, pages 2679--2713, 2025.

\bibitem[Tang et~al.(2025)Tang, Zhang, Hao, Wang, Wu, Wang, and
  Zhang]{tang2025affordgrasp}
Y.~Tang, S.~Zhang, X.~Hao, P.~Wang, J.~Wu, Z.~Wang, and S.~Zhang.
\newblock {AffordGrasp}: In-context affordance reasoning for open-vocabulary
  task-oriented grasping in clutter.
\newblock In \emph{2025 IEEE/RSJ International Conference on Intelligent Robots
  and Systems (IROS)}, pages 9433--9439, 2025.

\bibitem[Li et~al.(2025)Li, Wang, Qian, He, Wei, Wang, and Gong]{li2025dia}
J.~Li, S.~Wang, B.~Qian, Y.~He, X.~Wei, Q.~Wang, and Y.~Gong.
\newblock Dynamic integration of task-specific adapters for class incremental
  learning.
\newblock In \emph{Proceedings of the IEEE/CVF Conference on Computer Vision
  and Pattern Recognition (CVPR)}, pages 30545--30555, 2025.

\bibitem[Zhang et~al.(2026)Zhang, Wang, Tang, Wu, and Nie]{zhang2026sce}
S.~Zhang, S.~Wang, H.~Tang, J.~Wu, and L.~Nie.
\newblock Learning new tasks via reusable skills: Skill-compositional experts
  for embodied continual learning.
\newblock \emph{arXiv preprint arXiv:2606.15685}, 2026.

\bibitem[Zhao et~al.(2023)Zhao, Kumar, Levine, and Finn]{zhao2023act}
T.~Z. Zhao, V.~Kumar, S.~Levine, and C.~Finn.
\newblock Learning fine-grained bimanual manipulation with low-cost hardware.
\newblock In \emph{Proceedings of Robotics: Science and Systems}, 2023.

\bibitem[Chi et~al.(2023)Chi, Feng, Du, Xu, Cousineau, Burchfiel, and
  Song]{chi2023diffusion}
C.~Chi, S.~Feng, Y.~Du, Z.~Xu, E.~Cousineau, B.~C. Burchfiel, and S.~Song.
\newblock {Diffusion Policy}: Visuomotor policy learning via action diffusion.
\newblock In \emph{Proceedings of Robotics: Science and Systems}, 2023.

\bibitem[Oquab et~al.(2023)Oquab, Darcet, Moutakanni, Vo,
  et~al.]{oquab2023dinov2}
M.~Oquab, T.~Darcet, T.~Moutakanni, H.~Vo, et~al.
\newblock {DINOv2}: Learning robust visual features without supervision.
\newblock \emph{arXiv preprint arXiv:2304.07193}, 2023.

\bibitem[Peebles and Xie(2023)]{peebles2022dit}
W.~Peebles and S.~Xie.
\newblock Scalable diffusion models with transformers.
\newblock In \emph{Proceedings of the IEEE/CVF International Conference on
  Computer Vision (ICCV)}, pages 4195--4205, 2023.

\bibitem[Black et~al.(2025)Black, Brown, Darpinian, Dhabalia, Driess, Esmail,
  Equi, Finn, Fusai, Galliker, Ghosh, Groom, Hausman, ichter, Jakubczak, Jones,
  Ke, LeBlanc, Levine, Li-Bell, Mothukuri, Nair, Pertsch, Ren, Shi, Smith,
  Springenberg, Stachowicz, Tanner, Vuong, Walke, Walling, Wang, Yu, and
  Zhilinsky]{pi2025pi05}
K.~Black, N.~Brown, J.~Darpinian, K.~Dhabalia, D.~Driess, A.~Esmail, M.~R.
  Equi, C.~Finn, N.~Fusai, M.~Y. Galliker, D.~Ghosh, L.~Groom, K.~Hausman,
  b.~ichter, S.~Jakubczak, T.~Jones, L.~Ke, D.~LeBlanc, S.~Levine, A.~Li-Bell,
  M.~Mothukuri, S.~Nair, K.~Pertsch, A.~Z. Ren, L.~X. Shi, L.~Smith, J.~T.
  Springenberg, K.~Stachowicz, J.~Tanner, Q.~Vuong, H.~Walke, A.~Walling,
  H.~Wang, L.~Yu, and U.~Zhilinsky.
\newblock {$\pi_{0.5}$}: a vision-language-action model with open-world
  generalization.
\newblock In \emph{Proceedings of The 9th Conference on Robot Learning}, volume
  305, pages 17--40, 2025.

\bibitem[Hu et~al.(2021)Hu, Shen, Wallis, Allen-Zhu, Li, Wang, Wang, and
  Chen]{hu2021lora}
E.~J. Hu, Y.~Shen, P.~Wallis, Z.~Allen-Zhu, Y.~Li, S.~Wang, L.~Wang, and
  W.~Chen.
\newblock {LoRA}: Low-rank adaptation of large language models.
\newblock \emph{arXiv preprint arXiv:2106.09685}, 2021.

\bibitem[Ross et~al.(2011)Ross, Gordon, and Bagnell]{ross2011dagger}
S.~Ross, G.~Gordon, and D.~Bagnell.
\newblock A reduction of imitation learning and structured prediction to
  no-regret online learning.
\newblock In \emph{Proceedings of the Fourteenth International Conference on
  Artificial Intelligence and Statistics}, volume~15, pages 627--635, 2011.

\bibitem[{MetaFine Organizing
  Committee}(2026{\natexlab{a}})]{metafine_challenge}
{MetaFine Organizing Committee}.
\newblock {MetaFine Challenge}: Competition and evaluation protocol.
\newblock \url{https://github.com/Hiangx-robotics/MetaFine-Challenge_CoRL},
  2026{\natexlab{a}}.
\newblock COMPETITION.md and eval/README.md; accessed October 27, 2026.

\bibitem[{MetaFine Organizing
  Committee}(2026{\natexlab{b}})]{metafine_leaderboard}
{MetaFine Organizing Committee}.
\newblock {MetaFine Open Competition}: Final public leaderboard.
\newblock \url{https://robofinemani2026.github.io/leaderboard.html},
  2026{\natexlab{b}}.
\newblock Updated October 24, 2026; accessed October 27, 2026.

\bibitem[Black et~al.(2024)Black, Brown, Driess, Esmail, Equi, Finn, Fusai,
  Groom, Hausman, Ichter, Jakubczak, Jones, Ke, Levine, Li-Bell, Mothukuri,
  Nair, Pertsch, Shi, Tanner, Vuong, Walling, Wang, and
  Zhilinsky]{black2024pi0}
K.~Black, N.~Brown, D.~Driess, A.~Esmail, M.~Equi, C.~Finn, N.~Fusai, L.~Groom,
  K.~Hausman, B.~Ichter, S.~Jakubczak, T.~Jones, L.~Ke, S.~Levine, A.~Li-Bell,
  M.~Mothukuri, S.~Nair, K.~Pertsch, L.~X. Shi, J.~Tanner, Q.~Vuong,
  A.~Walling, H.~Wang, and U.~Zhilinsky.
\newblock {$\pi_0$}: A vision-language-action flow model for general robot
  control.
\newblock \emph{arXiv preprint arXiv:2410.24164}, 2024.

\end{thebibliography}
\fi

\end{document}